%% file: kdd2017.tex
\documentclass[sigconf]{acmart}

\usepackage{booktabs} 

\usepackage{algorithm}
\usepackage{algorithmic}
\usepackage[font=normalsize]{subfig}

\input{Definitions}

\setcopyright{none}

\acmYear{..\ ..\ .}
\acmDOI{XXXXXXXX}
\acmISBN{..\ ..\ .}
\acmConference[XXX'XX]{XX}{XXXX 20XX}{XXX} 
\acmPrice{XX.XX}

\begin{document}
\title[Grafting for Combinatorial Boolean Model using Frequent Itemset Mining]{Grafting for Combinatorial Boolean Model\\using Frequent Itemset Mining}

\author{Taito Lee}
\affiliation{%
  \institution{The University of Tokyo}
  \streetaddress{7-3-1, Hongo, Bunkyo}
  \city{Tokyo} 
  \country{Japan}
  \postcode{113-8656}
}
\email{ri.taito@ci.i.u-tokyo.ac.jp}

\author{Shin Matsushima}
\affiliation{%
  \institution{The University of Tokyo}
  \streetaddress{7-3-1, Hongo, Bunkyo}
  \city{Tokyo} 
  \country{Japan}
  \postcode{113-8656}
}
\email{shin_matsushima@mist.i.u-tokyo.ac.jp}

\author{Kenji Yamanishi}
\affiliation{%
  \institution{The University of Tokyo}
  \streetaddress{7-3-1, Hongo, Bunkyo}
  \city{Tokyo} 
  \country{Japan}
  \postcode{113-8656}
}
\email{yamanishi@mist.i.u-tokyo.ac.jp}

\renewcommand{\shortauthors}{T. Lee, S. Matsushima and K. Yamanishi}

\begin{abstract}

This paper introduces 
the {\em combinatorial Boolean model}~(CBM), which is defined as the class
of linear combinations of conjunctions of Boolean attributes.
This paper addresses the issue of learning CBM from labeled data. 
CBM is of high knowledge interoperability but  na\"{i}ve learning  of it requires exponentially large computation time with respect to data dimension and sample size.
To overcome this computational difficulty, we propose an  algorithm GRAB (GRAfting for Boolean datasets), which efficiently learns CBM within the $L_1$-regularized loss minimization framework.
The key idea of GRAB is to reduce the loss minimization problem to the weighted frequent itemset mining, in which frequent patterns are efficiently computable.
We employ benchmark datasets to empirically demonstrate that GRAB is 
effective in terms of computational efficiency, prediction accuracy and knowledge discovery.

\end{abstract}

%
%



\maketitle

\section{Introduction}
\label{sec:Introduction}

\subsection{Motivation}
We are concerned with learning classification/regression functions from labeled data.
We require here interpretability of acquired knowledge as well as high classification accuracy.
Under this requirement we focus on the class of linear prediction models of the following form:
\[f_{\bm{w}}(\bm{x})=\sum_{j=1}^dw_jx_j+w_{d+1},\]
 where $\bm{w}$ is a $(d+1)$-dimensional vector.
It takes as input a $d$-dimensional vector~$\bm{x}=(x_{1},\dots , x_{d})\in \mathbb{R}^d$ and outputs a
scalar value $y=f_{\bm{w}}(\bm{x})$. We call each $x_{j}$ an {\em attribute}.
This type of linear prediction functions is suitable for knowledge discovery in the sense that 
the attributes whose weights are large can be interpreted as more important {\em features} for 
classification or regression. 
Non-linear models such as the kernel method or the multi-layer neural networks may achieve higher classification accuracy, but have no such obvious interpretation of features.
Mind the
The power of  knowledge representation for linear predictor models is, however,  quite limited in the sense that an important feature is represented by a single attribute only.  
We are rather interested in a wider class of linear prediction models such that features may be represented by {\em combinations} of attributes.
Such a class have richer knowledge representations and may achieve higher prediction accuracy, if properly learned.
However, we may suffer from the computational difficulty in learning 
such a class since the total number of combinations of attributes is exponential in the dimension $d$.
There arises an issue of how we can learn such a class of rich knowledge representations with less computational demands.

The purpose of this paper is twofold.
The first is to propose an efficient algorithm that learns a class of linear predictors over all possible combination of binary attributes.
The second is to empirically demonstrate that the proposed algorithm efficiently produces  predictors having higher accuracy as well as better interpretability than competitive methods.

\subsection{Significance and Novelty}

The significance of this paper is summarized as follows:

{\em 1) Proposal of an efficient algorithm that learns a class of linear predictors over all conjunctions of attributes:}
We consider the class of linear predictors over all conjunctions of Boolean attributes.
We call this class the {\em combinatorial Boolean model}~(CBM).
It offers rich knowledge representation over the Boolean domain.
We consider the problem of learning CBM from labeled examples within the regularized  loss minimization framework. 
Since the size of CBM is exponential
in the data dimension, it is challenging to learn CBM as efficiently as possible.

We propose a novel algorithm called the {\em GRAfting for Boolean datasets} algorithm~(GRAB) to learn CBM efficiently.
It outputs a linear predictor in CBM in polynomial time
in sample size~$m$ and $|\mathcal{P}_\mathrm{all} |$,
where $\mathcal{P}_\mathrm{all}$ is the combinations of attributes which occur at least once in the input dataset.
The key idea of GRAB is to reduce the loss minimization problem  into the {\em frequent itemset mining}~(FIM) one.  
We use the {\em grafting algorithm}~\cite{Perkins03} designed for the large-scaled regularized loss minimization. We noticed that the grafting algorithm includes 
FIM as a sub-procedure. 
Meanwhile, there exists an algorithm for FIM \cite{Uno04}, which is
 able to efficiently find frequent itemsets making use of the monotonicity property of itemsets.
We successfully unified the efficient FIM algorithm into the grafting algorithm so that GRAB works very efficiently.

{\em 2) Empirical demonstration of validity of GRAB in terms of computational efficiency, prediction accuracy and knowledge interpretability:}
We employ benchmark datasets to demonstrate that GRAB is 
effective in terms of computational efficiency, prediction accuracy and knowledge intepretability.
We empirically show that GRAB achieves higher or almost comparable prediction accuracy with much less computational complexity than existing methods such as support vector classifiers with polynomial kernels and radial basis function kernels.
Further, we show that GRAB can acquire important knowledge in a comprehensive form.


\subsection{Related Work}
There are a lot of studies on learning interpretable knowledge representations such as decision trees, Boolean functions, etc.
Meanwhile, there are also many studies on uninterpretable but highly predictive knowledge representations such as support vector machines, neural networks, etc. We note that there also exist some studies that attempt to discover comprehensive knowledge using highly predictive models. Setiono et al.~\cite{Setiono02} proposed an algorithm that approximates
a trained neural network using combinations of linear predictors.
The resulting combinations of predictors are of high interpretability.

FIM  has been successful 
in several tasks emerging in machine learning such as clustering and boosting \citep{Saigo07,Tsuda06,Kudo04}.
One of most closely related work is \citep{Saigo07}, which utilized
FIM for boosting. They also considered a similar but not identical
model to CBM and worked on regression tasks on a biological context.

In the case of binary classification tasks,
learning of CBM  is related to learning Boolean functions,
especially disjunctive normal forms (DNF),
which has extensively been explored in the area of computational learning theory, e.g. \cite{Aizenstein95} and \cite{Bshouty95}.
However, CBM is different from DNF regarding to the two points: 
1. CBM takes a weighted sum of  conjunctions of attributes rather than disjunctive operations.
2. CBM includes all scholar valued functions: $\{0,1\}^{d}\rightarrow \mathbb{R}$.
Hence CBM can be thought of as a wider class of functions than DNF.
Learning CBM is significant in this sense.

The rest of this paper is organized as follows:
Section 2 proposes CBM.
Section 3 introduces the grafting algorithm for loss minimization.
Section 4 introduces the frequent itemset mining methodology.
Section 5 proposes our GRAB by combining the grafting algorithm with the frequent item mining.
Section 6 shows experimental results.
Section 7 gives concluding remarks.



\section{Combinatorial Boolean Model}
\label{sec:combinatorial_boolean_model}
This section introduces a class of combinatorial Boolean models.
Suppose that $\bm{x}\in \{0,1\}^d$, that is, each datum is represented by 
a Boolean valued vector. 
This assumption does not loose generality because 
when a datum is real-valued or integer-valued, we may transform it into a Boolean valued one by digitalizing it in some appropriate way (see Section 6).

Let $\mathcal{X}$ be the set of all attributes.
We define the combinatorial feature set $\Phi^{(d,k)}$ 
as a set of conjunctions of at most $k$ distinct attributes  chosen from $\mathcal{X}$.
For example, in case of $d=4, k=2$, $\Phi^{(d,k)}$ is given as follows:
\begin{align*}
\Phi^{(4,2)} = \{\top(\bm{x}), x_1, x_2, x_3, x_4, &x_1\wedge x_2, x_1\wedge x_3, x_1\wedge x_4, \\&
x_2\wedge x_3, x_2\wedge x_4, x_3\wedge x_4\},
\end{align*}
where $\top(\bm{x})\triangleq 1\ (\forall \bm{x}\in\{0,1\}^d)$ is the identity function.
Note that  $|\Phi^{(d,k)}|=\sum_{k'=0}^k \binom{d}{k'}$. Specifically,  $$|\Phi^{(d,d)}|=2^d.$$
When $d$ is fixed, we denote $\Phi^{(d,k)}$ as $\Phi^{(k)}$ in the discussion to follow.
We define a linear predictor associated with $\Phi^{(k)}$ by
\begin{align}\label{cbm}
f^{(k)}_{\bm{w}}(\bm{x})\triangleq\sum_{\phi\in\Phi^{(k)}}w_\phi \phi(\bm{x}),
\end{align}
where $\bm{w}=(w_{\phi})$ is a real-valued $|\Phi^{(k)}|$-dimensional parameter vector.
We call the class of all functions of the form (\ref{cbm}) the class of {\em combinatorial Boolean models}, which we abbreviate as CBM.
We call each $\phi \in \Phi ^{(k)}$ a {\em feature} and $k$ the {\em degree}.
The weight $w_{\phi}$ in $\bm{w}$ for $\phi$ represents the importance of the feature $\phi$. 
Hence CBM has good interpretability because  an important feature can be represented in a comprehensive form of a conjunction of attributes.

On the other hand, as for the complexity of CBM, the following proposition holds:
\begin{proposition}
Let $(\bm{x}_i,y_i)_{i=1}^m$ be the labeled examples,
where $\bm{x}_i\in \{0,1\}^d$ and $y_i\in\mathbb{R}$ for all $i=1,\ldots,m$.
We assume that $y_i=y_{i'}$ in the case of $\bm{x}_i=\bm{x}_{i'}$.
Then, for any $(\bm{x}_i,y_i)_{i=1}^m$, 
there exists $(\bm{w}_\phi)_{\Phi^{(d)}}$ that satisfies the following equation:
\begin{align*}
y_i=f^{(d)}_{\bm{w}}(\bm{x}_i)\ (i=1,\ldots,m).
\end{align*}
\end{proposition}
The proof is omitted but to appear in the full version.
This proposition shows that CBM is of high representability that CBM can fit any labeled examples.

Let us consider the problem of learning CBM. 
The purpose of learning is to estimate the parameter vector  $\bm{w}$ from the
given labeled examples $(\bm{x}_{1},y_{1}),\dots , (\bm{x}_{m},y_{m})$ where 
$y_{i}$ is the label corresponding to $\bm{x}_{i}$ and $m$ is the sample size. 
As a learning framework, we employ that of  {\em regularized loss minimization},  following \eqref{eq:ERM}.
In it the objective function for learning is given by
\begin{align}
 \label{eq:ERM}
 G(\bm{w}) \triangleq C \sum_{i=1}^m{ \ell \rbr{f_{\bm{w}}(\bm{x}_i ) , y_i}} + \Omega(\bm{w}),
\end{align}
where $\ell$ is a loss function, $\Omega$ is a regularizing function, and 
$C$ is a constant positive real number.

Our learning setting is closely related to loss
 minimization using polynomial kernels. 
In the case where $\bm{x}$ is a Boolean vector, 
 the loss minimization for CBM is analogous with 
that using  polynomial kernels~\cite{Shawe04}:
 \begin{align}
 k(\bm{x},\bm{x'})=(\bm{x}^\top\bm{x'}+ r)^l.
 \end{align}
 However, the weights for $\phi \in \Phi^{(k)}$s depend on their degrees or their numbers of conjunctions. Thus features for CBM are no-uniformly weighted while 
those for polynomial kernels are uniformly weighted.

From the standpoint of knowledge interpretability,
it is desired that large weights are assigned only to 
 a relatively small number of features $\phi\in\Phi^{(k)}$ after learning.
To this end we employ the framework of {\em sparse learning}~\cite{Rish14} by the 
$L_1$-regularizer $\Omega(\bm{w}) = \nbr{\bm{w}}_1$ in  \eqref{eq:ERM}.
We denote the objective function for this case 
 as $G^{(k)}(\bm{w})$: 
\begin{align*}
G^{(k)}(\bm{w})&\triangleq C \sum_{i=1}^m\ell(f^{(k)}_{\bm{w}}(\bm{x}_i),y_i)+\|\bm{w}\|_1 \\
& = C\sum_{i=1}^m\ell\left(\sum_{\phi\in\Phi^{(k)}}w_\phi \phi(\bm{x}_i),y_i\right)+\|\bm{w}\|_1.
\end{align*}

It is computationally difficult to learn CBM using existing standard techniques for loss minimization. 
In applying them, all of the values $\phi\rbr{\bm{x}_i}$ should be stored for all $\phi$ and $i$ beforehand.
It requires exponentially large memories with respect to the data dimension.
Further, it is also computationally expensive to solve a large-scale optimization problem associated with the loss minimization
since the total number of parameters is $\sum_{k'=0}^k \binom{d}{k'}$, which is at most $2^d$.
We show how to overcome this computational difficulty in the sections to follow.

\input{grafting}

\input{fim}

\input{proposed}
\input{experiments}

\section{Conclusion}

In this paper we proposed GRAB that is an algorithm for learning combinatorial boolean models~(CBM). 
The key idea of GRAB was to 
incorporate the techniques of frequent itemset mining with the grafting algorithm for regularized loss minimization.
We showed that GRAB was able to learn CBM more efficiently than the competitive methods such as the kernel methods.
We also showed that it achieved 
higher or comparable prediction accuracy than the competitive ones.
We also demonstrated that GRAB enabled us to discover knowledge in a form of conjunctions of attributes of given data.
This knowledge representation turned out to be very comprehensive.

The main reason for the efficiency of GRAB is that the monotonicity of itemsets, or boolean inputs, makes it easy to search over all possible features. 
Therefore, any other data structures having monotonicity (e.g. sequences, graphs, etc.) can also be incorporated with our methodology.

In this paper, we considered only the convex loss functions in order to avoid 
the algorithm from being trapped at local minima of the objective function. 
However, it is possible that GRAB works for non-convex losses as well as for convex ones.
It is worthwhile to note that GRAB works whenever the loss function is partially differentiable.
Hence, as future work, it is challenging to apply our methodology of GRAB into 
the efficient computation of multi-layer neural networks or other kinds of highly predictive machine learning models.

\bibliography{bib}
\bibliographystyle{ACM-Reference-Format}


\end{document}

%% file: grafting.tex
\section{Grafting Algorithm}
\label{sec:grafting}

In this section we introduce the grafting algorithm~\citep{Perkins03}.
Suppose that the data dimension is so large that there are many irrelevant attributes in the parameter vector $\bm{w}$. Then we may employ the $L_1$-regularizer as the penalty term $\Omega$ in (\ref{eq:ERM}), which is written as 
\begin{align}
\label{eq:L1}
 G(\bm{w}) = C \sum_{i=1}^m{ \ell \rbr{f_{\bm{w}}(\bm{x}_i ) , y_i}} + \nbr{\bm{w}}_1.
\end{align}
The grafting algorithm is designed so that it can solve the optimization problem of large-scale efficiently for such cases. 
The key idea of the grafting algorithm is to construct a set of active features by adding features incrementally.

\subsection*{Overall procedure}
In each iteration of the grafting algorithm, a (sub)gradient-based heuristics is employed to find the feature that seemingly improve the model most effectively and then to add it to the set of active features. At the $t$-th iteration, the grafting algorithm divides the set of all attributes of parameter vector $\bm{w}$ into two disjoint sets: $F^t$ and $Z^t\triangleq \neg F^t$. We call $w_j \in F^t$
{\em free weights}. $Z^t$ is constructed implicitly so that it always satisfies $w_j =0$ if $w_j \in Z^t$. 

The overall procedure of the grafting algorithm as follows:
First, it minimizes (\ref{eq:L1}) with respect to free weights, resulting in
\begin{align}
\label{eq:cond_weights_in_f}
 　\partial_{w_j}G \ni 0 
\end{align}
for $\forall j\in F^t$, where $\partial_{w_j}G $ is the subdifferential of $G$ with respect to $w_j$. 
Then, for $\forall j\in Z^t$, 
\begin{align}
  \label{eq:cond}
 　\partial_{w_j}G \not\ni 0 
\end{align}
is a necessary and sufficient condition for the value of the objective function to decrease by 
changing the $w_j$ (globally in case of convex $G$ and locally in general). 
Secondly, the grafting algorithm selects a parameter from $Z^t$ that is seemingly most effective for the objective function to decrease and adds it into $F^{t+1}$.
Then, $Z^{t+1}$ is also implicitly updated by removing the selected parameter, and the 
grafting algorithm iterates the procedure mentioned above.

\subsection*{Condition on effective parameters}
The subdifferential of the objective function with respect to  $w_j\in Z^t$
is calculated as
\begin{align}
\partial_{w_j}G 
\label{eq:subdiff}
&= \left\{c\ \middle| \frac{\partial L}{\partial w_j} - 1 \le c \le \frac{\partial L}{\partial w_j} + 1 \right\} ,
\end{align}
where $L(\bm{w})\triangleq\sum_{i=1}^m\ell(f_{\bm{w}}(\bm{x}_i),y_i)$.
Hence the condition~\eqref{eq:cond} is equivalent with 
\begin{align}
\label{eq:subdiff_cond}
\left|C\frac{\partial L}{\partial w_j}\right| > 1.
\end{align}
This implies that changing the value of $w_j$ from $0 $
will not decrease the objective function  if \eqref{eq:subdiff_cond} is not satisfied.
This is the main reason why $L_1$ regularization gives  a sparse
solution. It also leads to a stopping condition of the grafting algorithm as shown below.

\subsection*{Parameter selection}
We consider the problem of selecting a parameter to be moved from $Z^t$ to $F^t$.
We see from the above argument that $w_j\in Z^t$ satisfying
\eqref{eq:subdiff_cond} makes the  value of the objective function  decrease by 
changing its value from $0$. 
There may exist more than one candidates that
satisfy \eqref{eq:subdiff_cond}.
In that case, the grafting algorithm selects a parameter~$w_{\mathrm{best}}$ that makes the 
value of the objective function decrease most, by making use of the following 
gradient-based heuristics:
\begin{align}
\label{eq:grafting_weight_choice}
w_{\mathrm{best}}=\argmax_{w_j\in Z^t} \left|\frac{\partial L}{\partial w_j}\right|,
\end{align}
The derivatives of the objective function with respect to all parameters must be calculated to obtain 
the maximum in \eqref{eq:subdiff_cond}.
However, this na\"{i}ve method might be  computationally intractable when 
the data dimension and sample size are of large-scale.
To the best of  the authors' knowledge, there does not exist any  efficient method to solve this parameter selecting problem. 

\subsection*{Stopping condition}
If the condition \eqref{eq:subdiff_cond} is not satisfied
for all $w_j\in Z^t$, the value of the objective function is not decreased
by changing the value of $w_j\in Z^t$ from $0$ and the value of 
$w_j\in F^t$ from the current value, globally in case of convex $G$ and 
locally in general cases.
Therefore this condition can be used as a stopping condition for the grafting
algorithm. 
If the condition is fulfilled,  we may think that a local optimum (or the global optimum
in case of convex $G$) is achieved.

To summarize, the overall procedure is given in Algorithm~\ref{algo:grafting}.

\begin{algorithm}[!tb]
\caption{The grafting algorithm for $L_1$-regularized problem}
\label{algo:grafting}
\begin{algorithmic}[1]
\REQUIRE $G(\bm{w})=L(\bm{w})+\|\bm{w}\|_1$
\STATE $F^0\leftarrow\emptyset, Z^0\leftarrow \{w_j\}_{j=1}^{\mathrm{dim}(\bm{w})},t\gets 0$
\WHILE {$\max_{w_j \in Z} \left|\partial L/\partial w_j\right|>1$}
  \STATE $w_j\leftarrow \argmax_{w_j\in Z} \left|\partial L/\partial w_j\right|$
  \STATE $F^{t+1}\gets F^t \cup \{w_j\} $;\ $Z^{t+1}\gets Z^t \setminus \{w_j\} $
  \STATE $t \gets t+1$
  \STATE Optimize $G(\bm{w})$ with respect to $\forall w_j\in F^t$
\ENDWHILE
\end{algorithmic}
\end{algorithm}

%% file: fim.tex
\section{Frequent Itemset Mining}
\label{sec:fim}
\begin{figure*}
\begin{center}
\includegraphics[clip,width=2\columnwidth]{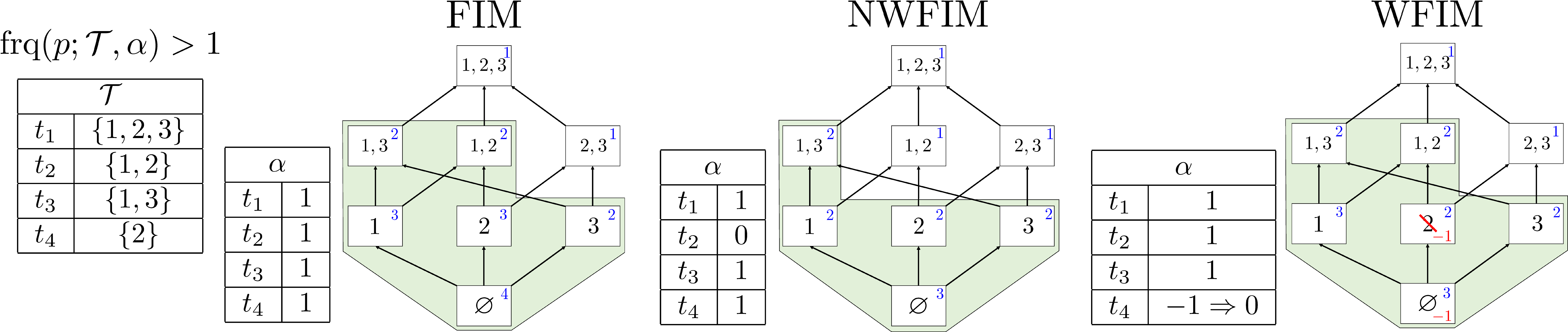}
\end{center}
\caption{Demonstration of frequent itemset mining (FIM) and its variants: in the standard setting of FIM, all the transaction are treated homogeneously 
and it searches all itemsets that appears in a given transaction database more than once (left panel).  Non-negatively weighted FIM (NWFIM)
deal with non-negative weights with respect to each transaction and find itemsets weighted frequency of which is more than 1 (middle panel).
Weighted FIM (WFIM) allows weights on transaction to be negative and loses the monotonicity of output (right panel).}
\label{fig:fim}
\end{figure*}
It is computationally difficult to find the best parameter according to \eqref{eq:grafting_weight_choice}. 
This is because  it requires computation of the gradient of loss over 
 all of the components of the parameter.
In order to overcome this difficulty, we employ the technique for frequent itemset mining~(FIM). In this section  we briefly review FIM.

\subsection*{Terminology}
A set of items $\mathcal{I}=\{1,\ldots,d\}$ is called the \emph{item base}. 
The set $\mathcal{T}=\{t_1,\ldots,t_m\}$ is called the {\it transaction database},
where each $t_i$ is a subset of $\mathcal{I}$.
Each element of  the transaction database  is called {\it a transaction}. Given a transaction database, an {\it occurrence set} of $p$, denoted by $\mathcal{T}(p)$,
is a set of all transactions that include $p$, i.e.,  
\begin{align}
\mathcal{T}(p) \triangleq \{t\in\mathcal{T}\ |\ t\supseteq p\}.
\end{align}
We refer to a subset of item base $p$ as an {\it itemset}. The cardinality of $\mathcal{T}(p)$ is called frequency, which is denoted as $\mathrm{frq}(p;\mathcal{T})$:
\begin{align}
\mathrm{frq}(p;\mathcal{T}) \triangleq \sum_{ t \in \mathcal{T}(p)} 1.
\end{align}
The simplest example of FIM problem is given as follows: For a given transaction database $\mathcal{T}$ and threshold $\theta$,  find $\mathcal{P}$, which is the set of all itemsets with a larger frequency than $\theta$, i.e.,
\begin{align}
\mathcal{P} = \{p\subseteq \mathcal{I}\ |\ \mathrm{frq}(p;\mathcal{T}) > \theta\}.
\end{align}

\subsection{Efficient Algorithms by Utilizing Monotonicity }
It is obvious that any subset of an itemset $p$ is included by a transaction $t$ when 
$p$ is included by $t$.
In other words, a kind of monotonicity holds in the following sense:
\begin{align}
\label{eq:monotonicity}
\mathcal{T}(p')\supseteq\mathcal{T}(p), \ \mathrm{ and } \ \mathrm{frq}(p';\mathcal{T})\geq\mathrm{frq}(p;\mathcal{T}),
\end{align}
for all $p' \subset p$.  
By making use of this property,
we can search all frequent itemsets by adding an item one by one from $\emptyset$.
The algorithm that performs this search in the breadth-first manner is the {\em apriori algorithm},
whereas the one that performs  this search in the depth-first manner is the {\em backtracking algorithm}.
The apriori algorithm was firstly proposed by \cite{Agrawal94}.
An FIM algorithm based on the backtraking algorithm was proposed in  
e.g., ~\cite{Zaki97} and~\cite{Bayardo98}.

The size of transaction database, denoted as $\nbr{\mathcal{T}}$, is defined by $\|\mathcal{T}\|  \triangleq \sum_{i=1}^m|t_i|$.
Time complexities for both of the apriori and backtracking algorithms are $\mathcal{O}(m\|\mathcal{T}\||\mathcal{P}|)$, which is called the {\em output-polynomial time}.
Hence, they are expected to run in practical time 
as long as $|\mathcal{P}|$ is small.

\subsection*{Extension to Weighted FIM}
Standard FIM methods handle transactions as if there were no difference in 
importance among transactions. 
However, there often appear such cases
that importance may differ one another depending on transactions.
Actually, the values of importance may be positive and negative.
If each transaction is allocated to a positive or negative label,
then we may be interested in discovering itemsets that frequently appear in positive transactions and
not frequently in negative ones.

In the setting where only the positive importance is available,
we can utilize the backtracking algorithm or the apriori algorithm.  
An efficient 
algorithm can be constructed, since the monotonicity of frequent itemsets still holds.
We define {\it weighted frequency} as follows:
\begin{align}
\mathrm{frq}(p;\mathcal{T},\alpha) \triangleq \sum_{ t \in \mathcal{T}(p)} \alpha_t.
\end{align}
for given $\alpha_t$ for $t\in \mathcal{T}$.
The same monotonicity as \eqref{eq:monotonicity} still holds for the weighted variant of frequencies.
We are then led to non-negatively-weighted FIM. 
It is to find all itemsets with larger weighted frequency than a given threshold.

In the setting where both positive and negative importance has to be dealt with,
the monotonicity  does not hold any longer.
Then any algorithm of output-polynomial time may not exist.
However, we may instead employ the following two-stage strategy:
Let the sets of positive and negative transactions in $\mathcal{T} $ be $\mathcal{T}_+ $ and $\mathcal{T}_- $, respectively.
In the first stage, ignoring transactions with negative importance, we obtain frequent itemsets such that 
\begin{align}
\mathcal{P}_+ = \{p\subseteq \mathcal{I}\ |\ \mathrm{frq}(p ; \mathcal{T}_+, \alpha )>\theta\}.
\end{align}
In the second stage,  for each itemset in $\mathcal{P}_+$ we check if the frequency $\mathrm{frq}(p ; \mathcal{T})$ 
is still larger than $\theta$. 
The first stage is executed using 
the aforementioned algorithms in time  $\mathcal{O}(m\|\mathcal{T}_+\||\mathcal{P_+}|)$ 
while the second stage is executed in time
 $\mathcal{O}(m\|\mathcal{T_-}\||\mathcal{P_+}|)$ by accessing all negative transactions for each itemset obtained in 
the first stage. Then the total computation time is $\mathcal{O}(m\|\mathcal{T}\||\mathcal{P}_+|)$. 

The three types of FIM: FIM, non-negatively weighted FIM (NWFIM), and weighted FIM (WFIM) are illustrated in Figure~\ref{fig:fim}.


A variant of WFIM can also be designed by restricting the itemsets  so that 
their sizes are at most $k$.
Then it will output $\mathcal{P}=\{p\subseteq \mathcal{I}\ |\ \mathrm{frq}(p;\mathcal{T},\alpha), |p|\leq k\}$.
It is realized by doing the breadth/depth-first search of frequent itemsets with at most $k$ depth.

%% file: proposed.tex
\section{Proposed Algorithm (GRAB) }
\label{sec:grab}

In this section we introduce our proposed algorithm for learning CBM, 
which we call GRAB (GRAfting for Boolean datasets).
As shown in Section~\ref{sec:combinatorial_boolean_model}, CBM has at most $2^d$ parameters. 
It is difficult to store all the values $\phi(\bm{x}_i)$ for all $\phi$ and $i$ from the viewpoint of space complexity. Further, it is difficult to 
solve this large-scale optimization problem from the viewpoint of time complexity.
However, we can overcome these difficulties by using the grafting algorithm in combination with the WFIM.
Let us consider to solve the optimization problem for learning CBM by means of the grafting algorithm.
It can overcome the space complexity issue because it does not require that all the possible features be stored.
On the other hand, a time complexity issue arises
in the process of finding a new feature in each iteration.

First note that the partial differential of the objective function~$G^{(k)}$ with respect to $w_\phi$ in case $w_\phi=0$ is given as follows:
\begin{align*}
\partial_{w_\phi}G 
&=
\bigg\{c\ \bigg|
C\sum_{i=1}^m
\left(\frac{\partial\ell}{\partial f^{(k)}(\bm{x}_i)}\phi(\bm{x}_i)\right) - 1 \le c, \\
&
\phantom{.....\bigg\{c\ \bigg|}
C\sum_{i=1}^m
\left(\frac{\partial\ell}{\partial f^{(k)}(\bm{x}_i)}\phi(\bm{x}_i)\right) + 1 \ge c \ \, \bigg\}.
\end{align*}
We see from \eqref{eq:subdiff_cond} that the value of the objective function is decreased by 
changing the value of $w_\phi$ if and only if
\begin{align}
\label{eq:grab_cond11}
\left|C\sum_{i=1}^m\left(\frac{\partial L}{\partial f^{(k)}(\bm{x}_i)}
\phi(\bm{x}_i)\right)\right| > 1.
\end{align}
Thus, the problem of finding the best feature is reduced
to finding $\phi \in \Phi^{(k)}$ satisfying \eqref{eq:grab_cond11}.
To solve this problem, we utilize the technique of WFIM. 
Note that the feature vector~$\bm{x}$ is an element of $\{0,1\}^d$.
Let us define a bijection~$T^{(d)}(\cdot)$ from ~$\{0,1\}^d$ to $2^{\{1,2,\ldots,d\}}$
and an injection~$P^{(k)}(\cdot)$ from $\Phi^{(k)}$ to $2^{\{1,2,\ldots,d\}}$
as follows:
\begin{align*}
T^{(d)}(\bm{x}) &\triangleq \{k\mid x_k=1\ (k=1,\ldots, d)\}, \\
P^{(k)}(\phi)&\triangleq\left\{ \begin{array}{lr}
\varnothing & (\phi = \top(\bm{x})), \\
\{i_1,\ldots,i_l\} & (\phi = x_{i_1}\wedge\ldots\wedge x_{i_l}).
\end{array}
\right.
\end{align*}
Using these functions, it holds that
\begin{align*}
\phi(\bm{x}) &= \left\{\begin{array}{lr}
1 & (T^{(d)}(\bm{x})\supset P^{(k)}(\phi)) ,\\
0 & (\text{otherwise.}).
\end{array}\right.
\end{align*}
We abbreviate $T^{(d)}(\bm{x}_i)$ and $P^{(k)}(\phi)$ as $t_i$ and $p_\phi$, respectively,
for each feature vector~$\bm{x}_i\ (i=1,\ldots,m)$ and $\phi\in\Phi^{(k)}$.
The left-hand side of \eqref{eq:grab_cond11} is rewritten as
\begin{align*}
\left|C\sum_{i=1}^m\left(\frac{\partial \ell}{\partial f^{(k)}(\bm{x}_i)}
\phi(\bm{x}_i)\right)\right| &= 
\left|C\sum_{\{i\ |\ \phi(\bm{x}_i)=1\}}\frac{\partial \ell}{\partial f^{(k)}(\bm{x}_i)}\right| \\
&=\left|C\sum_{\{i\ |\ t_i \supseteq\ p_\phi \}}\frac{\partial \ell}{\partial f^{(k)}(\bm{x}_i)}\right| \\
&=\left|C\sum_{\{i\ |\ t_i\in \mathcal{T}(p_\phi ) \}}\frac{\partial \ell}{\partial f^{(k)}(\bm{x}_i)}\right|,
\end{align*}
where $\mathcal{T}(p_\phi )$ is an occurence set of $p_\phi$ with respect to the transaction database 
$\mathcal{T}$ that regards $t_i$ as each transaction for $i=1,\ldots,m$.
Hence, we see \eqref{eq:grab_cond11} is rewritten as
\begin{align}
\mathrm{frq}(p_\phi ; \mathcal{T}, \alpha) > 1 
\label{eq:grab_cond_pos_neg} 
\quad \text{or} \quad
\mathrm{frq}(p_\phi ; \mathcal{T} , -\alpha) > 1,
\end{align}
where
\begin{align}
\label{eq:transaction_weights}
\alpha_{t_i} = C\frac{\partial \ell}{\partial f^{(k)}(\bm{x}_i)} .
\end{align}
In other words,
we can obtain a set of all $p_\phi$s that satisfy \eqref{eq:grab_cond_pos_neg}
by employing WFIM for the transaction database~$\mathcal{T}$
and the transaction weights as in~\eqref{eq:transaction_weights}.
According to the original form of grafting algorithm, only one parameter is newly added to free weights
by \eqref{eq:grafting_weight_choice} at each iteration.
However, knowing that \eqref{eq:grab_cond11} is a sufficient and necessary condition for the 
objective function to decrease, we may select more than one  parameters such as 
the top-K frequent parameters: $\{w_{\phi _{i}}\}_{i=1}^{K}$ where $p_{\phi _{i}}$ is the $i$-th most frequent itemset in $\mathcal{P}$. 
This method is more efficient than the parameter estimation step of the grafting algorithm
in the case where the parameter selection procedure of WFIM requires much computation time.

The overall flow of GRAB algorithm is given in Algorithm~\ref{algo:grab}.
The time complexity of GRAB algorithm is evaluated as follows: 
At each step of GRAB, we perform WFIM and optimization of $G^{(k)}(\bm{w})$.
Since time complexity for the optimization largely depends on the loss function and the employed solver,
we only consider time complexity for WFIM step.
WFIM takes much more time than the optimization in many cases.
As discussed at Section~\ref{sec:fim}, the computation time of WFIM is $\mathcal{O}\rbr{m\nbr{\mathcal{T}}|\mathcal{P}_+|}$.
Here, it obviously holds that $\mathcal{P}_+\subseteq \mathcal{P}_\mathrm{all}$,
where $\mathcal{P}_\mathrm{all}\triangleq\{p\subseteq\mathcal{I}\ |\ \mathrm{frq}\rbr{p;\mathcal{T}}>0\}$.
The total computation time for WFIM  in GRAB is eventually evaluated as follows:
$\mathcal{O}\rbr{mT_1\nbr{\mathcal{T}}|\mathcal{P}_\mathrm{all}|}$, 
where $T_1$ is total number of iteration.

\begin{algorithm}[!tb]
\caption{GRAB algorithm}
\label{algo:grab}
\begin{algorithmic}[1]
\INPUT $k\ge 0, \{\bm{x}_i ,y_i\}_{i=1,\ldots,m}, C>0$
\REQUIRE $
\mathcal{T} = \cbr{t_i}_{i=1,\ldots,m}
$
\STATE $t_i \gets T^{(d)} (\bm{x}_i), \ \alpha_{t_i} \gets C\frac{\partial \ell}{\partial f^{(k)}(\bm{x}_i)}  \quad ( i =1,\ldots,m) $
\STATE $\mathcal{P} \gets \cbr{\ p\  | \  \mathrm{frq}(p_\phi ; \mathcal{T}, \alpha) > 1 \ \text{or} \ \mathrm{frq}(p_\phi ; \mathcal{T} , -\alpha) > 1, |p| \le k}$
\STATE $F\leftarrow\varnothing, Z\leftarrow \{w_\phi\}_{\phi\in\Phi^{(k)}}$
\WHILE {$\mathcal{P}\not=\varnothing$}
  \FOR{$j=1,\ldots,K$ }
  \STATE Pick $\phi$ such that $p_{\phi}$ is the $j$-th most frequent itemset in $\mathcal{P}$
  \STATE Move $w_{\phi}$ to $F$ from $Z$.
  \ENDFOR
  \STATE Optimize $G^{(k)}(\bm{w})$ w.r.t. $w_\phi\in F$
  \STATE $\alpha_{t_i} \gets C\frac{\partial \ell}{\partial f^{(k)}(\bm{x}_i)} $
  \STATE $\mathcal{P} \gets \cbr{\ p\  | \  \mathrm{frq}(p_\phi ; \mathcal{T}, \alpha) > 1 \ \text{or} \ \mathrm{frq}(p_\phi ; \mathcal{T} , -\alpha) > 1, |p| \le k}$
\ENDWHILE
\OUTPUT $\{w_\phi\}_{\phi\in F}$
\end{algorithmic}
\end{algorithm}

\subsection*{Implementation Issue}

We describe details of  implementation for acceleration
and termination of GRAB.
\paragraph{Dynamic threshold control for the acceleration of WFIM}
We have already shown that the time complexity of WFIM is 
$\Ocal(m\|\mathcal{T}\||\mathcal{P}_+|)$.
Hence WFIM terminates faster if the threshold $\theta$ is larger and $|\mathcal{P}_+|$ is smaller.
Further, it is desired to find the top-$K$ frequent itemsets without extracting all the features that satisfy \eqref{eq:grab_cond11}. 
To this end we first execute WFIM by setting the threshold $\theta = 2^M$ with $M=10$,
and decrement $M$ by $1$ until we obtain $K$ itemsets or it becomes $\theta=2^0=1$.
When WFIM is called next time, we use the same value of $M$ as that the previously used one. 

\paragraph{Incomplete Termination of WFIM}
When the total number of outputs of WFIM is too large,
we terminate WFIM after extracting $100K$ frequent itemsets and
select the top-$K$ frequent itemsets among them.
In such cases, it may not be guaranteed that the top-$K$ frequent itemsets are 
selected.
However, as the selection of a new feature based on \eqref{eq:grafting_weight_choice} is already a heuristic, 
we do not expect that it significantly deteriorate the performance of the optimization step.

\paragraph{Stopping condition}
We employ the following stopping condition: 
First we define the suboptimality of 
a solution as follows:
\begin{align*}
V^{(t)}\triangleq\sum_{\phi\in\Phi^{(k)}}v_\phi^{(t)} 
	   =\sum_{w_\phi\in F^t}v_\phi^{(t)} + \sum_{w_\phi\in Z^t}v_\phi^{(t)},
\end{align*}
where 
\begin{align*}
v_\phi^{(t)}&=\begin{cases}
C\left|\frac{\partial L}{\partial w_\phi^{(t)}} + \mathrm{sgn}\rbr{w_\phi^{(t)}}\right| & (w_\phi^{(t)}\ne 0) \\
\max\left(C\left|\frac{\partial L}{\partial w_\phi^{(t)}}\right| - 1, 0\right) & (w_\phi^{(t)}=0).
\end{cases}\end{align*}

For a given tolerance $\varepsilon>0$, 
we terminate the algorithm when $V^{(t)} < \varepsilon V^{(0)}$ is satisfied.
The first term of $V^{(t)}$ is computed using the parameter obtained by the optimization step. 
The second term is computed by summing over features obtained by WFIM.
$V^{(0)}$ may not be able to be computed exactly as long as we employ the heuristics for acceleration.
We underestimate $V^{(0)}$ by computing the summand in the second term
among the obtained features. This makes the stopping condition stricter and the proposed algorithm run longer.
In our experiments, we set $\varepsilon=0.01$.

%% file: experiments.tex
\section{Experiments}
\label{sec:experiments}

We implemented our algorithm GRAB
\footnote{All of our experimental codes are available at 
\url{https://www.dropbox.com/sh/fjjmwkiwt509368/AABR7L2at0vQc4xMld6_WCGoa?dl=0}.}
in C and C++
to 
evaluate its computational complexity, prediction accuracy, and knowledge interpretability.
In order to select new features to be added, we used Linear time Closed itemset Miner (LCM) by ~\cite{Uno03,Uno04,Uno05}, which is one of backtracking-based FIM  algorithms. All experiments below are executed by Linux (CentOS 6.4) machines 
with 96GB memory and Intel(R) Xeon(R) Processor X5690 @ 3.47GHz.
We restricted the computation time to within one day
and experiments with overtime were taken as time-outs.

\subsection{Evaluation of Computation Time}
In order to evaluate the computational efficiency of GRAB, 
we employed benchmark datasets to compare GRAB with other methods.
We investigated how effective the grafting algorithm and WFIM were respectively.
To this end, as methods for comparison, we employed the following two combinations:
\begin{enumerate}
\item{{\em Expansion + $L_1$-regularized logistic regression:}\\}
In this combination, we first expand a dataset so that  all possible features are employed to express $\phi\in\Phi^{(k)}$ ({\em expansion}).
We then learn combinatorial linear models on the expanded data set within the $L_1$-regularized logistic regression framework. As a solver, we used LIBLINEAR\cite{liblinear}.
\item{{\em Grafting + Na\"{\i}ve feature selection:}\\}
In this combination we employ the grating algorithm, without combing it with WFIM. 
In it we calculate \eqref{eq:grafting_weight_choice} by searching exhaustively over  the set of all features~$\phi\in\Phi^{(k)}$.
\end{enumerate}

In our experiment, we used the logistic loss function $\ell(f(\bm{x}), y)\triangleq -\log\rbr{1/\rbr{1+\exp\rbr{-yf(\bm{x}}}}$ 
as a loss function for learning CBM.
As a benchmark data set, we used a1a dataset~\cite{Platt99},
which has 32,561 records and 123 attributes.
We conducted learning CBM with $C=0.1, 1$ and $k=1,2,3,4,5,6,\infty$.  
The results on computation time are shown in Table~\ref{tab:time}.

\begin{table*}[htbp]
\caption{
Comparison of computation time (sec) on a1a dataset:
the bracketed values at \textit{Expansion+$L_1$-regularized Logistic Regression}
mean the elapsed times to solve $L_1$-regularized logistic regression.
}
\label{tab:time}
\begin{center}
\begin{tabular}{|r||r|r|r|r|l|l|} \hline
           & \multicolumn{2}{|c|}{GRAB} & \multicolumn{2}{|c|}{Grafting + Na\"{i}ve feature selection} & \multicolumn{2}{|c|}{Expansion + $L_1$-regularized Logistic Regression} \\ \hline
           & \multicolumn{1}{|c|}{$C=0.1$} & \multicolumn{1}{|c|}{$C=1$} & \multicolumn{1}{|c|}{$C=0.1$}  & \multicolumn{1}{|c|}{$C=1$} & \multicolumn{1}{|c|}{$C=0.1$} & \multicolumn{1}{|c|}{$C=1$} \\ \hline
$k=1$      & 2.1 & 1.7 & \qquad\qquad\quad 5.3 & 4.0 & \qquad\quad \textbf{3.0E-1}\ (1.4E-1) & \textbf{2.4E-2}\ (8.5E-2) \\ \hline
$k=2$      & 4.5 & 5.8E+1 & 3.2E+2 & 2.9E+2 & \textbf{3.8}\ (3.0E-1) & \textbf{3.8}\ (3.2E-1) \\ \hline
$k=3$      & \textbf{8.7} & \textbf{2.1E+1} & 1.3E+4 & 1.2E+4 & 5.5E+1\ (9.8E-1) & 5.5E+1\ (1.1) \\ \hline
$k=4$      & \textbf{1.3E+1} & \textbf{4.1E+1} & $> 1$ day        & $> 1$ day        & 4.7E+2\ (1.4) & 4.7E+2\ (1.9) \\ \hline
$k=5$      & \textbf{1.5E+1} & \textbf{9.2E+1} & $> 1$ day        & $> 1$ day        & 2.3E+3\ (5.1) & 2.3E+3\ (5.2) \\ \hline
$k=6$      & \textbf{1.7E+1} & \textbf{1.4E+2} & $> 1$ day        & $> 1$ day        & 7.6E+3\ (1.3E+1) & 7.6E+3\ (1.1E+1) \\ \hline
$k=\infty$ & \textbf{2.3E+1} & \textbf{2.9E+2}  &$ > 1$ day        & $> 1$ day        & 5.6E+4\ (9.2E+1) & 5.6E+4\ (9.1E+1) \\ \hline
\end{tabular}
\end{center}
\end{table*}

We first observe that the computation time for GRAB is significantly smaller than that of Grafting+Na\"{\i}ve feature selection.
It implies that WFIM was so efficient that it could  select the features even when
exhaustive search is impractical.
LIBLINEAR could solve the optimization problem more quickly than GRAB. 
However, the expansion step took much longer time than the whole procedure of GRAB.
Therefore we conclude that GRAB is more efficient than the compared methods. We consider the main reason of the efficiency is that GRAB could extract features selectively. 

\subsection{Evaluation of prediction accuracy}

In order to evaluate GRAB's prediction performance,
we conducted experiments for the following benchmark datasets:
\begin{enumerate}
\item
\textbf{a1a}\footnotemark[1]~\cite{Platt99}: It has 32,561 records and 123 attributes for each data.
This dataset is obtained by preprocessing UCI Adult dataset~\cite{UCI},
which has 6 continuous and 8 categorical features.
We divided the total data into 30,000 training data and 2,561 test data.
\item
\textbf{cod-rna}\footnotemark[1]~\cite{Uzilov06}:
It is a dataset for the problem of detecting non-coding RNAs from attributes of base sequences.
Since all of  the 8 attributes are real-valued,
we transformed  each attribute into a binary one as follows:
We divided the interval from the minimum value to the maximum value into 30 cells of equal length.
We expressed each attribute in a binary form by indicating which section its value fell into.
Thus, a binary dataset with 240 attributes was obtained.
Each data in it was specified by  8 non-zero attributes.
We used 300,000 data for training and 31,152 for test.
\item
\textbf{covtype.binary}\footnotemark[1]~\cite{Collobert02}:
It is a dataset obtained by transforming UCI Covertype dataset~\cite{UCI}
with multi labels into the one with binary labels.
The original Covertype dataset has 581,012 instances with 54 cartographic attributes,
and 10 of which are quantitative ones and the rest are binary.
We binarized all the quantitative attributes in the same way as cod-rna dataset.
We used 500,000 data for training and 81,012 for test.
\end{enumerate}
\footnotetext[1]{
Datasets are available on the LIBSVM Data website at
https://www.csie.ntu.edu.tw/~cjlin/libsvmtools/datasets/binary.html.
}

As methods for comparison, we employed {\em support vector classification} (SVC)~\cite{Bishop06}
with two types of kernel functions: polynomial kernel (Poly)~\cite{Shawe04}
and radial basis function kernel (RBF)~\cite{Shawe04},
both of which have extensively been used for the purpose of classification.
The polynomial kernel is defined as $k(\bm{x},\bm{x}')=\rbr{\gamma \bm{x}^\top\bm{x}'+\rbr{1-\gamma}}^d$,
where $d\in\mathbb{N}_+$ and $\gamma\in (0,1]$ are hyper parameters,
and radial basis function kernel is defined as $k(\bm{x},\bm{x}')=\exp\rbr{\frac{1}{2\gamma}\|\bm{x}-\bm{x}'\|^2_2}$,
where $\gamma>0$ is a hyper parameter.

We employed the logistic loss  and the $L_2$-hinge loss as loss functions for learning CBM.
The $L_2$-hinge loss function is defined as:
\begin{align*}
\ell(f(\bm{x}), y)\triangleq \begin{cases}
\rbr{1-yf(\bm{x})}^2 & {\rm if}\ 1-yf(\bm{x})>0 ,\\
0 & \text{otherwise.}
\end{cases}
\end{align*}
Note that the WFIM procedure in GRAB can be accelerated
by using the $L_2$-hinge loss.
This is because $\partial \ell(\bm{x}_i,y_i)/\partial f^{(k)}(\bm{x})$,
which is the $i$-th transaction weight of WFIM,
is $0$ for all $i$ satisfying $1-y_if(\bm{x}_i)\leq 0$
and we can remove the corresponding data from the transaction database.

For each of the benchmark datasets,
we compared prediction accuracies of the following algorithms:
\begin{enumerate}
\item Polynomial kernel SVC with $d=1$ and $3$,
\item RBF kernel SVC,
\item Logistic loss GRAB with $k=1,3$ and $\infty$, 
\item $L_2$-hinge loss GRAB with $k=1,3$ and $\infty$.
\end{enumerate}
For each of the algorithms,  
after learning models with $C=10^{-3},$ $10^{-2},$$\ldots,$ $10^3$ for GRAB
and with $C=10^{-3},$ $10^{-2},$$\ldots,$ $10^3$ and $\gamma=0.1,0.2,\ldots,1$ for SVC from training data,
we selected the model with the highest accuracy on test data.
In order to conduct SVCs,
we made use of scikit-learn~\cite{sklearn}, which is a machine learning library written in Python.
Note that it is not unfair to compare computation times of GRAB with them,
since the SVCs in scikit-learn call LIBSVM~\cite{LIBSVM},
which is written in C and C++.

Table~\ref{tab:accuracy} shows the accuracies of the respective algorithms.
From the results on a1a dataset, we see that GRAB with $k=1$ is 
almost comparable to Poly with $d=1$,
while the former finished training more than 100 times faster than the latter.
This implies that linear models are accurate enough for the prediction with a1a dataset.
The time complexity of logistic loss GRAB with $k=3$ 
was more than 10 times higher than polynomial kernel SVC with $d=3$.

From the results on cod-rna dataset shown in Table~\ref{tab:accuracy}\subref{tab:accuracy_cod-rna}, 
we observe that the accuracies of SVC on the binarized dataset are much greater than those on the raw one.
The reason for this may be that
cod-rna dataset becomes almost linearly separable by means of binarization.
$L_2$-hinge loss GRAB with  $k=\infty$ is more than $1\%$ more accurate than that with $k=1$. 
This implies that combinatorial features are effective for the prediction with cod-rna dataset.
Poly with $d=3$ on the binarized data, which achieved the highest accuracy among the SVC family, was almost comparable to $L_2$-hinge loss GRAB with $k=\infty$, 
which achieved the  highest accuracy among GRAB family.
However, the computation time of the best in GRABs was less than one hundredth of that of the best in SVCs.
This is because binarized cod-rna dataset was of high sparsity in the sense that the number of nonzero elements per record was just eight out of 240,
and WFIM was  executed very rapidly.

From the results on covtype.binary dataset shown in  Table~\ref{tab:accuracy}\subref{tab:accuracy_covtype.binary}, 
we see that  binarization contributed to the improvement of  accuracies of SVCs.
This was similar to the case of cod-rna.
We also see that combinatorial features greatly contributed to the prediction with covtype.binary dataset. In terms of computation time, GRAB outperformed SVC significantly. GRAB finished within two hours while SVC on the binarized dataset did not finish in one day.


\begin{table*}[htbp]
\caption{
Comparison of accuracy (\%) on benchmark datasets:
the terms \textit{Raw} and \textit{Binarized} mean that
we used the original and the binarized dataset, respectively,
as a training data.
The row \textit{Time} shows the elapsed time (sec) of each method
when the number of training data is the maximum.
}
\label{tab:accuracy}
\begin{center}
\subfloat[a1a \label{tab:accuracy_a1a}]{
\begin{tabular}{|r||r|r|r|r|r|r|r|r|r|} \hline
 & \multicolumn{2}{|c|}{Poly} & \multicolumn{1}{|c|}{RBF} & \multicolumn{3}{|c|}{GRAB (logistic loss)} & \multicolumn{3}{|c|}{GRAB ($L_2$-hinge loss)} \\ \hline
\#Data & $d=1$ & $d=3$ & & $k=1$ & $k=3$ & $k=\infty$ & $k=1$ & $k=3$ & $k=\infty$ \\ \hline
7,500 & 85.2 & 85.3 & 85.2 & 84.9 & 84.7 & 84.9 & 84.9 & \textbf{85.3} & 84.5 \\ \hline
30,000 & 85.4 & 85.3 & 85.2 & 85.0 & 85.5 & 85.5 & 85.1 & \textbf{85.5} & 84.3 \\ \hline\hline
Time (sec) & 2.3E+2 & 1.0E+2 & 1.4E+2 & 1.4 & 6.9 & 1.7E+1 & 1.2 & 1.4E+1 & 5.0 \\ \hline
\end{tabular}
}

\subfloat[cod-rna \label{tab:accuracy_cod-rna}]{
\begin{tabular}{|r||r|r|r|r|r|r|r|r|r|r|r|r|} \hline
Dataset & \multicolumn{3}{|c|}{Raw} & \multicolumn{9}{|c|}{Binarized} \\ \hline
 & \multicolumn{2}{|c|}{Poly} & \multicolumn{1}{|c|}{RBF} & \multicolumn{2}{|c|}{Poly} & \multicolumn{1}{|c|}{RBF} & \multicolumn{3}{|c|}{GRAB (logistic loss)} & \multicolumn{3}{|c|}{GRAB ($L_2$-hinge loss)} \\ \hline
\#Data & $d=1$ & $d=3$ & & $d=1$ & $d=3$ & & $k=1$ & $k=3$ & $k=\infty$ & $k=1$ & $k=3$ & $k=\infty$ \\ \hline
50,000 & 88.9 & 88.8 & 89.3 & 95.6 & \textbf{96.0} & 96.0 & 95.5 & 95.9 & 95.9 & 95.5 & 96.0 & 96.0 \\ \hline
100,000 & 89.0 & 88.9 & 90.1 & 95.7 & 96.3 & \textbf{96.3} & 95.6 & 96.0 & 96.0 & 95.6 & 96.3 & 96.3 \\ \hline
300,000 & 89.4 & 89.0 & 93.0 & 95.7 & 96.6 & 96.4 & 95.6 & 96.5 & 96.5 & 95.6 & 96.6 & \textbf{96.7} \\ \hline\hline
Time (sec) & 3.0E+3 & 3.8E+3 & 5.3E+3 & 8.0E+3 & 1.4E+4 & 1.1E+4 & 6.4E+1 & 2.1E+2 & 1.1E+2 & 1.5E+1 & 8.8E+1 & 1.1E+2 \\ \hline
\end{tabular}
}

\subfloat[covtype.binary \label{tab:accuracy_covtype.binary}]{
\begin{tabular}{|r||r|r|r|r|r|r|r|r|r|r|r|r|} \hline
Dataset & \multicolumn{3}{|c|}{Raw} & \multicolumn{9}{|c|}{Binarized} \\ \hline
 & \multicolumn{2}{|c|}{Poly} & \multicolumn{1}{|c|}{RBF} & \multicolumn{2}{|c|}{Poly} & \multicolumn{1}{|c|}{RBF} & \multicolumn{3}{|c|}{GRAB (logistic loss)} & \multicolumn{3}{|c|}{GRAB ($L_2$-hinge loss)} \\ \hline
\#Data & $d=1$ & $d=3$ & & $d=1$ & $d=3$ & & $k=1$ & $k=3$ & $k=\infty$ & $k=1$ & $k=3$ & $k=\infty$ \\ \hline
50,000 & 76.0 & 82.0 & 86.1 & 74.4 & 86.8 & \textbf{87.6} & 76.9 & 86.7 & 86.9 & 76.9 & 86.5 & 86.6 \\ \hline
100,000 & 76.1 & 82.4 & 87.0 & 74.1 & 90.6 & \textbf{91.2} & 77.0 & 90.8 & 90.9 & 77.0 & 90.4 & 90.6 \\ \hline
500,000 & 76.0 & 81.0 & 84.8 & - & - & - & 76.9 & 94.1 & 94.3 & 77.0 & 94.3 & \textbf{94.7} \\ \hline\hline
Time (sec) & 3.0E+4 & 2.3E+4 & 3.1E+4 & $> 1$ day & $> 1$ day & $> 1$ day & 6.6E+1 & 3.7E+3 & 4.7E+3 & 3.8E+1 & 2.6E+3 & 3.6E+3 \\ \hline
\end{tabular}
}
\end{center}
\end{table*}

\subsection{Evaluation of interpretability}
We can conduct knowledge acquisition from the learning results for GRAB as follows:
Look at the predictor learned by GRAB.
In it the features whose corresponding weights have large absolute values can be interpreted as important ones.
This is because such features greatly affect the prediction results.
Hence, by simply extracting large-valued features,  we can acquire knowledge about what features are important for prediction.
Specifically, features used in GRAB are very comprehensive since they are represented as conjunctions of attributes.
This implies that GRAB is of high interpretability of the acquired knowledge.

Below we show the results on knowledge discovery for a1a dataset~\cite{Platt99}.
This dataset was obtained by binarizing the adult dataset, which was extracted from 
the census bureau database.
The a1a dataset has been used as a benchmark for classification of people with more than \$50,000 annual income or the others, on the basis of their attributes.

Table~\ref{tab:a1a_knowledge} shows weights with large absolute values,
which were obtained by GRAB with $L_2$-hinge loss and $k=3$ on a1a dataset.
The features having large absolute values of weight are listed in a descending order.
This list itself is of high intepretability and represents knowledge acquired from the dataset. 

The feature with the largest absolute value of  weight is (Not in family), which has a negative gain for classification.
The feature with the secondly largest absolute value of weight is 
(Not in family)\&(No monetary capital losses), which has a positive gain for classification. 
It is interesting to see that an identical attribute (Not in family) can contribute to both of the positive and negative gains for classification. 
This suggests the importance of conjunctions of attributes in knowledge discovery.

The feature  with the 6-th largest absolute value of weight is 
(large education number), which has a positive gain for classification. 
But its conjunction with (Other service) \& (Middle average hours per week worked) has a negative gain (the 9-th feature) while its conjunction with (Prof specialty)\&(Have monetary capital gains) has a positive gain (the 10-th feature).
Hence making combinations of attributes is really informative for deepening knowledge discovery.

\begin{table*}
\caption{
Weights with top-10 large absolute values obtained from a1a dataset.
}
\label{tab:a1a_knowledge}
\begin{tabular}{|r|c|} \hline
Weight & Combination of Attributes \\ \hline
-0.581 & (Not in family) \\ \hline
0.544 & (Not in family) \& (No monetary capital losses) \\ \hline
0.455 & (Self-employment (not inc.)) \& (Unmarried) \& (Have monetary capital gains) \\ \hline
-0.432 & (Female) \& (No monetary capital gains) \& (No monetary capital losses) \\ \hline
0.390 & (Low final weight) \& (Asian Pacific Islander) \& (From Japan) \\ \hline
0.353 & (Large education number) \\ \hline
0.325 & (Married AF spouse) \& (No monetary capital gains) \& (From United States) \\ \hline
0.320 & (Married civ spouse) \& (No monetary capital losses) \& (From Canada) \\ \hline
-0.291 & (Large education number) \& (Other service) \& (Middle average hours per week worked) \\ \hline
0.272 & (Slightly larger education number) \& (Prof specialty) \& (Have monetary capital gains) \\ \hline
\end{tabular}
\end{table*}

%% file: kdd2017.bbl

\begin{thebibliography}{00}


\ifx \showCODEN    \undefined \def \showCODEN     #1{\unskip}     \fi
\ifx \showDOI      \undefined \def \showDOI       #1{{\tt DOI:}\penalty0{#1}\ }
  \fi
\ifx \showISBNx    \undefined \def \showISBNx     #1{\unskip}     \fi
\ifx \showISBNxiii \undefined \def \showISBNxiii  #1{\unskip}     \fi
\ifx \showISSN     \undefined \def \showISSN      #1{\unskip}     \fi
\ifx \showLCCN     \undefined \def \showLCCN      #1{\unskip}     \fi
\ifx \shownote     \undefined \def \shownote      #1{#1}          \fi
\ifx \showarticletitle \undefined \def \showarticletitle #1{#1}   \fi
\ifx \showURL      \undefined \def \showURL       #1{#1}          \fi
\providecommand\bibfield[2]{#2}
\providecommand\bibinfo[2]{#2}
\providecommand\natexlab[1]{#1}

\bibitem[\protect\citeauthoryear{Agrawal and Srikant}{Agrawal and
  Srikant}{1994}]%
        {Agrawal94}
\bibfield{author}{\bibinfo{person}{Rakesh Agrawal} {and}
  \bibinfo{person}{Ramakrishnan Srikant}.} \bibinfo{year}{1994}\natexlab{}.
\newblock \showarticletitle{Fast algorithms for mining association rules}. In
  \bibinfo{booktitle}{{\em 20th int. conf. very large data bases, VLDB}},
  \bibinfo{volume}{Vol. 1215}. \bibinfo{pages}{487--499}.
\newblock


\bibitem[\protect\citeauthoryear{Aizenstein and Pitt}{Aizenstein and
  Pitt}{1995}]%
        {Aizenstein95}
\bibfield{author}{\bibinfo{person}{Howard Aizenstein} {and}
  \bibinfo{person}{Leonard Pitt}.} \bibinfo{year}{1995}\natexlab{}.
\newblock \showarticletitle{On the Learnability of Disjunctive Normal Form
  Formulas}.
\newblock \bibinfo{journal}{{\em Machine Learning\/}} \bibinfo{volume}{{19}, 3}
  (\bibinfo{year}{1995}), \bibinfo{pages}{183--208}.
\newblock


\bibitem[\protect\citeauthoryear{Andrew V.~Uzilov and Mathews}{Andrew V.~Uzilov
  and Mathews}{2006}]%
        {Uzilov06}
\bibfield{author}{\bibinfo{person}{Joshua M.~Keegan Andrew V.~Uzilov} {and}
  \bibinfo{person}{David~H. Mathews}.} \bibinfo{year}{2006}\natexlab{}.
\newblock \showarticletitle{Detection of non-coding RNAs on the basis of
  predicted secondary structure formation free energy change}.
\newblock \bibinfo{journal}{{\em BMC Bioinf.\/}} \bibinfo{volume}{{7}, 1}
  (\bibinfo{year}{2006}), \bibinfo{pages}{173}.
\newblock


\bibitem[\protect\citeauthoryear{Bayardo~Jr}{Bayardo~Jr}{1998}]%
        {Bayardo98}
\bibfield{author}{\bibinfo{person}{Roberto~J Bayardo~Jr}.}
  \bibinfo{year}{1998}\natexlab{}.
\newblock \showarticletitle{Efficiently mining long patterns from databases}.
\newblock \bibinfo{journal}{{\em ACM Sigmod Record\/}} \bibinfo{volume}{{27},
  2} (\bibinfo{year}{1998}), \bibinfo{pages}{85--93}.
\newblock


\bibitem[\protect\citeauthoryear{Bishop}{Bishop}{2006}]%
        {Bishop06}
\bibfield{author}{\bibinfo{person}{Christopher~M. Bishop}.}
  \bibinfo{year}{2006}\natexlab{}.
\newblock \bibinfo{booktitle}{{\em Pattern Recognition and Machine Learning
  (Information Science and Statistics)}}.
\newblock Springer-Verlag New York, Inc., Secaucus, NJ, USA.
\newblock


\bibitem[\protect\citeauthoryear{Bshouty}{Bshouty}{1995}]%
        {Bshouty95}
\bibfield{author}{\bibinfo{person}{Nader~H. Bshouty}.}
  \bibinfo{year}{1995}\natexlab{}.
\newblock \showarticletitle{Exact Learning Boolean Functions via the Monotone
  Theory}.
\newblock \bibinfo{journal}{{\em Information and Computation\/}}
  \bibinfo{volume}{{123}, 1} (\bibinfo{year}{1995}), \bibinfo{pages}{146--153}.
\newblock


\bibitem[\protect\citeauthoryear{Chang and Lin}{Chang and Lin}{2011}]%
        {LIBSVM}
\bibfield{author}{\bibinfo{person}{Chih-Chung Chang} {and}
  \bibinfo{person}{Chih-Jen Lin}.} \bibinfo{year}{2011}\natexlab{}.
\newblock \showarticletitle{{LIBSVM}: A library for support vector machines}.
\newblock \bibinfo{journal}{{\em ACM Transactions on Intelligent Systems and
  Technology\/}}  \bibinfo{volume}{2} (\bibinfo{year}{2011}),
  \bibinfo{pages}{27:1--27:27}.
\newblock
Issue 3.
\newblock
\shownote{Software available at
  \url{http://www.csie.ntu.edu.tw/~cjlin/libsvm}.}


\bibitem[\protect\citeauthoryear{Collobert, Bengio, and Bengio}{Collobert
  et~al\mbox{.}}{2002}]%
        {Collobert02}
\bibfield{author}{\bibinfo{person}{Ronan Collobert}, \bibinfo{person}{Samy
  Bengio}, {and} \bibinfo{person}{Yoshua Bengio}.}
  \bibinfo{year}{2002}\natexlab{}.
\newblock \showarticletitle{A Parallel Mixture of SVMs for Very Large Scale
  Problems}.
\newblock \bibinfo{journal}{{\em Neural compt.\/}} \bibinfo{volume}{{14}, 5}
  (\bibinfo{year}{2002}), \bibinfo{pages}{1105--1114}.
\newblock


\bibitem[\protect\citeauthoryear{Fan, Chang, Hsieh, Wang, and Lin}{Fan
  et~al\mbox{.}}{2008}]%
        {liblinear}
\bibfield{author}{\bibinfo{person}{Rong-En Fan}, \bibinfo{person}{Kai-Wei
  Chang}, \bibinfo{person}{Cho-Jui Hsieh}, \bibinfo{person}{Xiang-Rui Wang},
  {and} \bibinfo{person}{Chih-Jen Lin}.} \bibinfo{year}{2008}\natexlab{}.
\newblock \showarticletitle{LIBLINEAR: A library for large linear
  classification}.
\newblock \bibinfo{journal}{{\em Journal of machine learning research\/}}
  \bibinfo{volume}{{9}, Aug} (\bibinfo{year}{2008}),
  \bibinfo{pages}{1871--1874}.
\newblock


\bibitem[\protect\citeauthoryear{Kudo, Maeda, and Matsumoto}{Kudo
  et~al\mbox{.}}{2004}]%
        {Kudo04}
\bibfield{author}{\bibinfo{person}{Taku Kudo}, \bibinfo{person}{Eisaku Maeda},
  {and} \bibinfo{person}{Yuji Matsumoto}.} \bibinfo{year}{2004}\natexlab{}.
\newblock \showarticletitle{An application of boosting to graph
  classification.}. In \bibinfo{booktitle}{{\em NIPS}},
  \bibinfo{volume}{Vol.~17}. \bibinfo{pages}{729--736}.
\newblock


\bibitem[\protect\citeauthoryear{Lichman}{Lichman}{2013}]%
        {UCI}
\bibfield{author}{\bibinfo{person}{M. Lichman}.}
  \bibinfo{year}{2013}\natexlab{}.
\newblock \bibinfo{title}{{UCI} Machine Learning Repository}.
\newblock   (\bibinfo{year}{2013}).
\newblock
\showURL{%
\url{http://archive.ics.uci.edu/ml}}


\bibitem[\protect\citeauthoryear{Pedregosa, Varoquaux, Gramfort, Michel,
  Thirion, Grisel, Blondel, Prettenhofer, Weiss, Dubourg, Vanderplas, Passos,
  Cournapeau, Brucher, Perrot, and Duchesnay}{Pedregosa et~al\mbox{.}}{2011}]%
        {sklearn}
\bibfield{author}{\bibinfo{person}{F. Pedregosa}, \bibinfo{person}{G.
  Varoquaux}, \bibinfo{person}{A. Gramfort}, \bibinfo{person}{V. Michel},
  \bibinfo{person}{B. Thirion}, \bibinfo{person}{O. Grisel},
  \bibinfo{person}{M. Blondel}, \bibinfo{person}{P. Prettenhofer},
  \bibinfo{person}{R. Weiss}, \bibinfo{person}{V. Dubourg}, \bibinfo{person}{J.
  Vanderplas}, \bibinfo{person}{A. Passos}, \bibinfo{person}{D. Cournapeau},
  \bibinfo{person}{M. Brucher}, \bibinfo{person}{M. Perrot}, {and}
  \bibinfo{person}{E. Duchesnay}.} \bibinfo{year}{2011}\natexlab{}.
\newblock \showarticletitle{Scikit-learn: Machine Learning in {P}ython}.
\newblock \bibinfo{journal}{{\em Journal of Machine Learning Research\/}}
  \bibinfo{volume}{12} (\bibinfo{year}{2011}), \bibinfo{pages}{2825--2830}.
\newblock


\bibitem[\protect\citeauthoryear{Perkins, Lacker, and Theiler}{Perkins
  et~al\mbox{.}}{2003}]%
        {Perkins03}
\bibfield{author}{\bibinfo{person}{Simon Perkins}, \bibinfo{person}{Kevin
  Lacker}, {and} \bibinfo{person}{James Theiler}.}
  \bibinfo{year}{2003}\natexlab{}.
\newblock \showarticletitle{Grafting: Fast, incremental feature selection by
  gradient descent in function space}.
\newblock \bibinfo{journal}{{\em Journal of machine learning research\/}}
  \bibinfo{volume}{{3}, Mar} (\bibinfo{year}{2003}),
  \bibinfo{pages}{1333--1356}.
\newblock


\bibitem[\protect\citeauthoryear{Platt}{Platt}{1999}]%
        {Platt99}
\bibfield{author}{\bibinfo{person}{John~C. Platt}.}
  \bibinfo{year}{1999}\natexlab{}.
\newblock \showarticletitle{Advances in Kernel Methods}.
\newblock MIT Press, Cambridge, MA, USA, Chapter Fast Training of Support
  Vector Machines Using Sequential Minimal Optimization,
  \bibinfo{pages}{185--208}.
\newblock
\showISBNx{0-262-19416-3}


\bibitem[\protect\citeauthoryear{Rish and Grabarnik}{Rish and
  Grabarnik}{2014}]%
        {Rish14}
\bibfield{author}{\bibinfo{person}{Irina Rish} {and} \bibinfo{person}{Genady
  Grabarnik}.} \bibinfo{year}{2014}\natexlab{}.
\newblock \bibinfo{booktitle}{{\em Sparse Modeling: Theory, Algorithms, and
  Applications\/} (1st ed.)}.
\newblock CRC Press, Inc., Boca Raton, FL, USA.
\newblock


\bibitem[\protect\citeauthoryear{Saigo, Uno, and Tsuda}{Saigo
  et~al\mbox{.}}{2007}]%
        {Saigo07}
\bibfield{author}{\bibinfo{person}{Hiroto Saigo}, \bibinfo{person}{Takeaki
  Uno}, {and} \bibinfo{person}{Koji Tsuda}.} \bibinfo{year}{2007}\natexlab{}.
\newblock \showarticletitle{Mining complex genotypic features for predicting
  HIV-1 drug resistance}.
\newblock \bibinfo{journal}{{\em Bioinf.\/}} \bibinfo{volume}{{23}, 18}
  (\bibinfo{year}{2007}), \bibinfo{pages}{2455--2462}.
\newblock


\bibitem[\protect\citeauthoryear{Setiono, Leow, and Zurada}{Setiono
  et~al\mbox{.}}{2002}]%
        {Setiono02}
\bibfield{author}{\bibinfo{person}{Rudy Setiono}, \bibinfo{person}{Wee~Kheng
  Leow}, {and} \bibinfo{person}{Jacek~M. Zurada}.}
  \bibinfo{year}{2002}\natexlab{}.
\newblock \showarticletitle{Extraction of Rules From Artificial Neural Networks
  for Nonlinear Regression}.
\newblock \bibinfo{journal}{{\em IEEE Transactions on Neural Networks\/}}
  \bibinfo{volume}{{13}, 3} (\bibinfo{date}{May} \bibinfo{year}{2002}),
  \bibinfo{pages}{564--577}.
\newblock


\bibitem[\protect\citeauthoryear{Shawe-Taylor and Cristianini}{Shawe-Taylor and
  Cristianini}{2004}]%
        {Shawe04}
\bibfield{author}{\bibinfo{person}{John Shawe-Taylor} {and}
  \bibinfo{person}{Nello Cristianini}.} \bibinfo{year}{2004}\natexlab{}.
\newblock \bibinfo{booktitle}{{\em Kernel Methods for Pattern Analysis}}.
\newblock Cambridge University Press, New York, NY, USA.
\newblock


\bibitem[\protect\citeauthoryear{Tsuda and Kudo}{Tsuda and Kudo}{2006}]%
        {Tsuda06}
\bibfield{author}{\bibinfo{person}{Koji Tsuda} {and} \bibinfo{person}{Taku
  Kudo}.} \bibinfo{year}{2006}\natexlab{}.
\newblock \showarticletitle{Clustering graphs by weighted substructure mining}.
  In \bibinfo{booktitle}{{\em Proceedings of the 23rd international conference
  on Machine learning}}. ACM, \bibinfo{pages}{953--960}.
\newblock


\bibitem[\protect\citeauthoryear{Uno, Asai, Uchida, and Arimura}{Uno
  et~al\mbox{.}}{2003}]%
        {Uno03}
\bibfield{author}{\bibinfo{person}{Takeaki Uno}, \bibinfo{person}{Tatsuya
  Asai}, \bibinfo{person}{Yuzo Uchida}, {and} \bibinfo{person}{Hiroki
  Arimura}.} \bibinfo{year}{2003}\natexlab{}.
\newblock \showarticletitle{LCM: An Efficient Algorithm for Enumerating
  Frequent Closed Item Sets.}. In \bibinfo{booktitle}{{\em FIMI}},
  \bibinfo{volume}{Vol.~90}. Citeseer.
\newblock


\bibitem[\protect\citeauthoryear{Uno, Kiyomi, and Arimura}{Uno
  et~al\mbox{.}}{2004}]%
        {Uno04}
\bibfield{author}{\bibinfo{person}{Takeaki Uno}, \bibinfo{person}{Masashi
  Kiyomi}, {and} \bibinfo{person}{Hiroki Arimura}.}
  \bibinfo{year}{2004}\natexlab{}.
\newblock \showarticletitle{LCM ver. 2: Efficient mining algorithms for
  frequent/closed/maximal itemsets}. In \bibinfo{booktitle}{{\em FIMI}},
  \bibinfo{volume}{Vol. 126}.
\newblock


\bibitem[\protect\citeauthoryear{Uno, Kiyomi, and Arimura}{Uno
  et~al\mbox{.}}{2005}]%
        {Uno05}
\bibfield{author}{\bibinfo{person}{Takeaki Uno}, \bibinfo{person}{Masashi
  Kiyomi}, {and} \bibinfo{person}{Hiroki Arimura}.}
  \bibinfo{year}{2005}\natexlab{}.
\newblock \showarticletitle{LCM ver. 3: collaboration of array, bitmap and
  prefix tree for frequent itemset mining}. In \bibinfo{booktitle}{{\em
  Proceedings of the 1st international workshop on open source data mining:
  frequent pattern mining implementations}}. ACM, \bibinfo{pages}{77--86}.
\newblock


\bibitem[\protect\citeauthoryear{Zaki, Parthasarathy, Ogihara, and Li}{Zaki
  et~al\mbox{.}}{1997}]%
        {Zaki97}
\bibfield{author}{\bibinfo{person}{Mohammed~Javeed Zaki},
  \bibinfo{person}{Srinivasan Parthasarathy}, \bibinfo{person}{Mitsunori
  Ogihara}, {and} \bibinfo{person}{Wei Li}.} \bibinfo{year}{1997}\natexlab{}.
\newblock \showarticletitle{New Algorithms for Fast Discovery of Association
  Rules.}. In \bibinfo{booktitle}{{\em Proc. of KDD}},
  \bibinfo{volume}{Vol.~97}. \bibinfo{pages}{283--286}.
\newblock


\end{thebibliography}
